\documentclass[pmlr]{jmlr}

\usepackage{longtable}

\usepackage{booktabs}
\usepackage[load-configurations=version-1]{siunitx} 
\usepackage{caption}
\usepackage{multirow}
\usepackage{xcolor}
\usepackage{colortbl}
\definecolor{mygray}{gray}{0.8}

\usepackage[T1]{fontenc}

\theorembodyfont{\upshape}
\theoremheaderfont{\scshape}
\theorempostheader{:}
\theoremsep{\newline}

\newcount\Comments   
\usepackage{color}
\definecolor{purple}{rgb}{1,0,1}
\definecolor{orange}{rgb}{.85,.25,0}
\newcommand{\kibitz}[2]{\ifnum\Comments=1\textcolor{#1}{#2}\fi}
\newcommand{\joyce}[1]{\kibitz{purple}      {[Joyce: #1]}}

\jmlrvolume{106}
\jmlryear{2019}
\jmlrworkshop{Machine Learning for Healthcare}

\title[MT-DTI]{Self-Attention Based Molecule Representation for Predicting Drug-Target Interaction}

 \author{\Name{Bonggun Shin$^{\S \dag}$}    \Email{bonggun.shin@emory.edu}\\
         \Name{Sungsoo Park$^{\dag}$}   \Email{sspark@deargen.me}\\
         \Name{Keunsoo Kang$^{\ddag \dag}$}   \Email{kangk1204@dankook.ac.kr}\\
         \Name{Joyce C. Ho$^{\S}$}     \Email{joyce.c.ho@emory.edu} \\
         $^{\S}$ \addr Department of Computer Science, Emory University, Atlanta, GA, USA \\
         $^{\dag}$ \addr Deargen Inc., Seoul, South Korea \\
         $^{\ddag}$ \addr Department of Microbiology, Dankook University, Cheonan, South Korea
 }

\begin{document}

\maketitle

\begin{abstract}
Predicting drug-target interactions (DTI) is an essential part of the drug discovery process, which is an expensive process in terms of time and cost.
Therefore, reducing DTI cost could lead to reduced healthcare costs for a patient. 
In addition, a precisely learned molecule representation in a DTI model could contribute to developing personalized medicine, 
which will help many patient cohorts. In this paper, we propose a new molecule representation based on the self-attention mechanism, 
and a new DTI model using our molecule representation.
The experiments show that our DTI model outperforms the state of the art by up to 4.9\% points in terms of area under the precision-recall curve.
Moreover, a study using the DrugBank database proves that our model effectively lists all known drugs targeting a specific cancer biomarker in the top-30 candidate list.

\end{abstract}

\section{Introduction}
Many diseases are caused by abnormal protein levels, therefore, 
a drug is designed to target particular proteins. 
However, a drug may not work well for a decent portion of patients, because an
individual's response to a drug varies depending on the genetic inheritance~\citep{wang2008pharmacogenomics}.
Unfortunately, pharmaceutical companies focus only on a majority cohort of patients as drug discovery is an expensive process.
The reduction of the cost of the drug discovery process will not only lead to drugs costing less, resulting in reduced healthcare costs for a patient but can also allow companies to develop personalized drugs based on genetics.

Among the many parts of the drug discovery process, predicting drug-target interactions (DTI) is an essential one.
DTI is difficult and costly as experimental assays not only take significant time but are expensive. 
Furthermore, only less than 10\% of the proposed DTIs are accepted as new drugs~\citep{he2017simboost}. 
Therefore, \textit{in silico} (performed on a computer) DTI predictions are much demanded 
since it can expedite the drug development process by systemically suggesting a new set of candidate molecules promptly, which can save time and reduce the cost of the whole process by up to 43\%~\citep{dimasi2016innovation}.

In response to this demand, three types of \textit{in silico} DTI prediction methods have been proposed in the literature: molecular docking, similarity-based, and deep learning-based models.
Molecular docking~\citep{trott2010autodock, luo2016molecular} is a simulation-based method using the 3D structured features of molecules and proteins. 
Although it can provide an intuitive visual interpretation, it is difficult to obtain a 3D structure of a feature and cannot scale to large datasets.
To mitigate these problems, two similarity-based methods, KronRLS~\citep{pahikkala2014toward} and SimBoost~\citep{he2017simboost} have been proposed 
using efficient machine learning methods. 
However, using a similarity matrix has two downsides. Firstly, feature representation is limited in the similarity space, thereby ignoring the rich information embedded in the molecule sequence. For example, if a brand new molecule is tested, the model will represent it using relatively unrelated (dissimilar) molecules, which would make the prediction inaccurate. Secondly, it necessitates the calculation of the similarity matrix which can limit the maximum number of molecules in the training process.
To overcome these limitations, a deep learning-based DTI model, DeepDTA~\citep{ozturk2018deepdta}, was proposed. It is an end-to-end convolutional neural network (CNN)-based model that eliminates the need for feature engineering. 
The model automatically finds useful features from a raw molecule and protein sequence.
Its success has been demonstrated on two publicly available DTI benchmarks.
Although this work illustrated the potential of a deep learning-based model, there are several areas for improvement:
\begin{itemize}
    \item CNNs can't model potential relationships among distant atoms in a raw molecule sequence. 
    For example, with three layers of CNNs each with a filter size of 12, the model can capture associations in atoms up to 35 distances in a sequence.
    We posit that the recently proposed self-attention mechanism can be used to capture any relationship among  atoms in a sequence, and thereby provide a better molecule relationship
    \item The one-hot encoding used to represent each molecule fails to take advantage of existing chemical structure knowledge. An abundance of chemical compounds are available in the PubChem database~\citep{10.1093/nar/gky1033}, from which we can extract useful chemical structures for pre-training the molecule representation network.
    \item Fine-tuning is a type of transfer learning where weights trained from one network can be transferred to another so that the weights can be adjusted to the new dataset. Thus, we can transfer the weights learned from the PubChem database to our DTI model. This will help our model to use the learned knowledge of a chemical structure while tailoring it to predicting DTI interactions.
\end{itemize}

With these observations, we propose a new deep DTI model, Molecule Transformer DTI (MT-DTI), based on a new molecule representation.
We use a self-attention mechanism to learn the high-dimensional structure of a molecule from a given raw sequence.
Our self-attention mechanism, Molecular Transformer (MT), is pre-trained on publicly available chemical compounds (PubChem database) to learn the complex structure of a molecule. 
This pre-training is important, because most datasets available for DTI training has only 2000 molecules, while the data for pre-training (PubChem database) contains 97 \textit{millions} of molecules.
Although it does not contain interaction data but just  molecules, 
our MT is able to learn a chemical structure from it, which will be effectively utilized when transferred to MT-DTI (our model).
Therefore, we transfer this trained molecule representation to our DTI model so that it can be fine-tuned with a DTI dataset. 
The proposed DTI model is evaluated on two well-known benchmark DTI datasets, Kiba~\citep{tang2014making} and Davis~\citep{davis2011comprehensive}, and outperforms the current state of the art (SOTA) model by 4.9\% points for Kiba and 1.6\% points for Davis in terms of area under the precision-recall curve.
Additionally, we demonstrate the usefulness of our trained model using a known drug list targeting a specific protein. 
The trained model generates all FDA approved drugs with high rankings in the drug candidate lists.
The demonstrated effectiveness of the proposed model can help reduce the cost of drug discovery.
Furthermore, precise molecule representation can enable drugs to be designed for specific genotypes and potentially enable personalized medicine.

\paragraph{Technical Significance}
We propose a novel molecule representation, adapting the self-attention mechanism that was recently proposed in Natural Language Process (NLP) literature.
This is inspired by the idea that understanding a molecule sequence for a chemist is analogous to understanding a language for a person. 
We introduce a new way to train the molecule representation model to fit the DTI problem using an existing corpus to achieve a more robust representation.
With this (pre)trained molecule representation, we fine-tune the proposed DTI model and achieve new SOTA performances on two public DTI benchmarks.\footnote{The demo is publicly available at\\: \url{https://mt-dti.deargendev.me/}}

\paragraph{Clinical Relevance}
With our new model, we can potentially lower medication costs for patients, which can help make drugs more affordable and help patients be more adherent. In addition, this can serve as the stepping stone for designing personalized medication. Through the proper representation of molecules and proteins, we can better understand the properties of patients that make a drug helpful or not~\citep{quinn2017molecular}.

\section{Methods}
\label{sec:methods}
\begin{figure}
	\centering
	\begin{minipage}[b]{0.8\textwidth}
		\includegraphics[width=\textwidth]{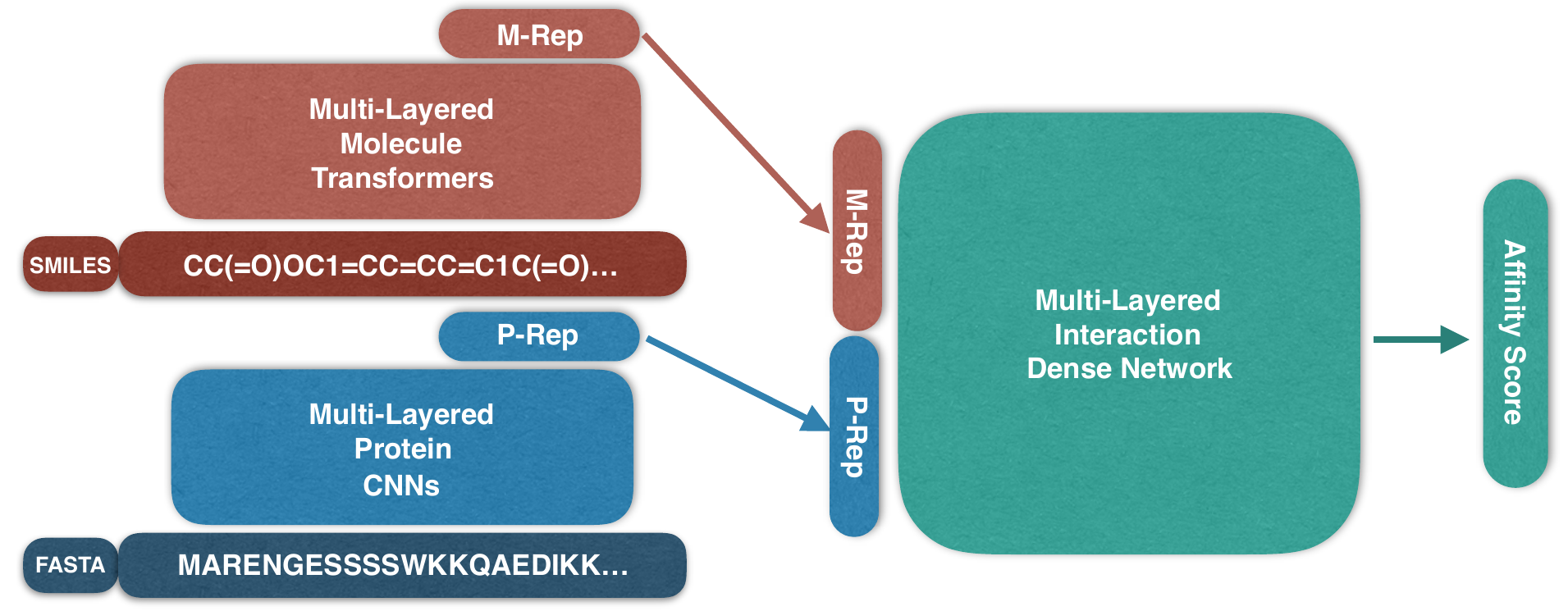}
		\caption{The Proposed DTI Model Architecture.
		Inputs are molecule (SMILES)  and protein (FASTA) and
		the regression output is the affinity score between these two inputs.
		}
		\label{fig:overall}
	\end{minipage}
\end{figure}

\begin{figure}
	\begin{minipage}[b]{0.32\textwidth}
		\includegraphics[width=0.95\textwidth]{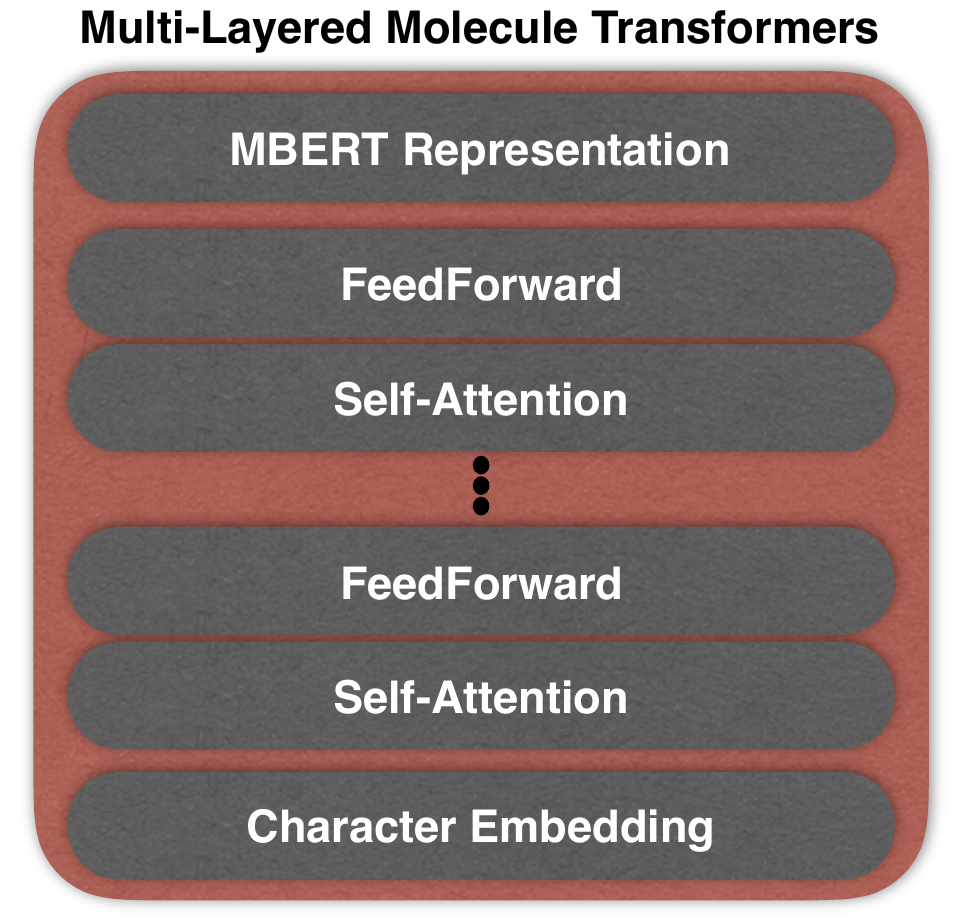}
		\caption{Molecule Trans.}
		\label{fig:moleule_transformer}
	\end{minipage}
	\begin{minipage}[b]{0.32\textwidth}
		\includegraphics[width=0.95\textwidth]{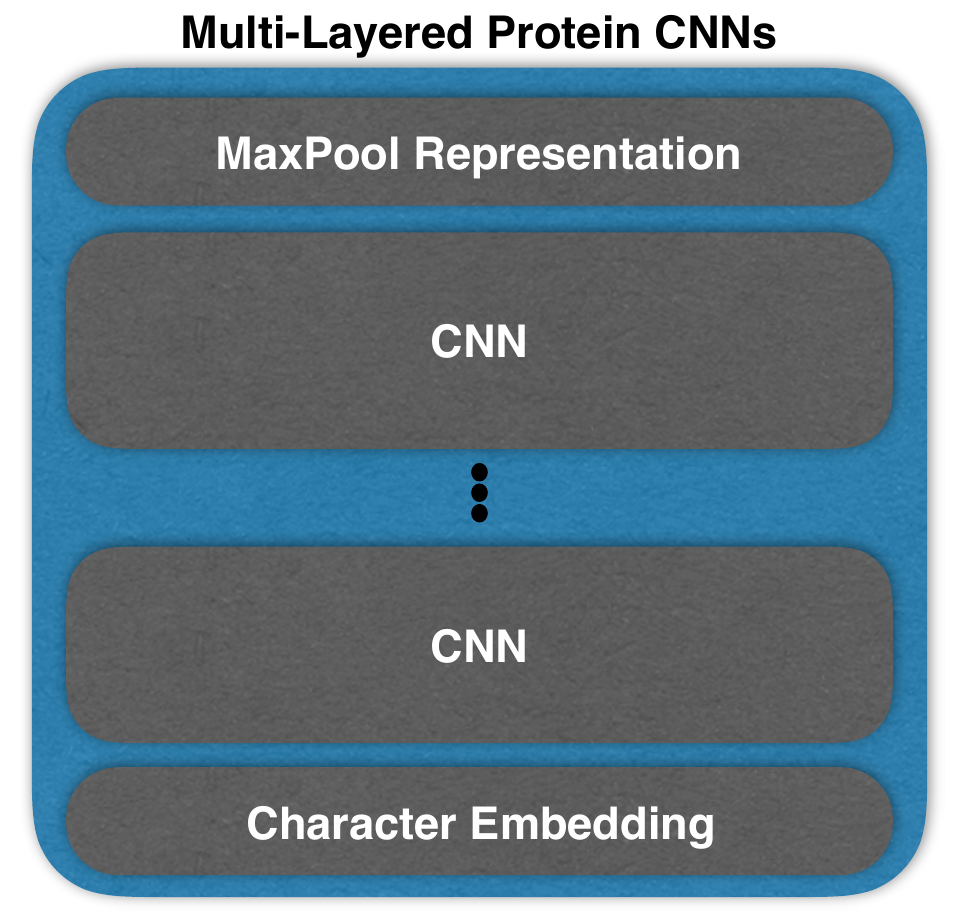}
		\caption{Protein CNNs.}
		\label{fig:protein_cnn}
	\end{minipage}
	\begin{minipage}[b]{0.32\textwidth}
		\includegraphics[width=0.95\textwidth]{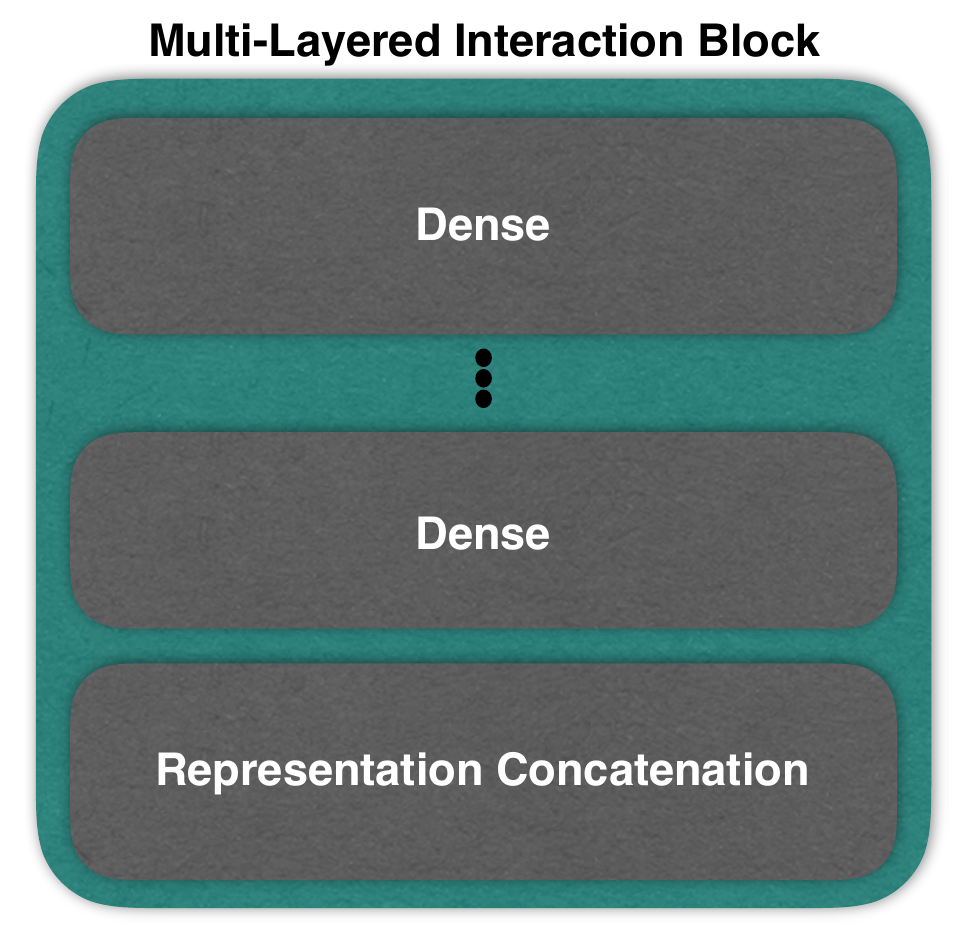}
		\caption{Interaction Denses.}
		\label{fig:interaction_dense}
	\end{minipage}
\end{figure}

We introduce a new drug-target interaction (DTI) model and a new molecule representation in this section.
The basic motivation of the proposed model is that the structure of molecule sequences is shown to be very similar to the structure of 
natural language sentences in that contextual and structural information of atoms are important when understanding the characteristics of a molecule~\citep{jastrzkebski2016learning}.
Specifically, each atom interacts with not only neighboring atoms but also long distant one in a  simplified molecular-input line-entry system (SMILES) sequence, a notation that encodes the molecular structure of chemicals.
However, the current SOTA method using CNNs can't relate long distance atoms when representing a molecule.
We overcome this using the self-attention mechanism.
We first describe the proposed MT-DTI model architecture (Figure~\ref{fig:overall}) with input and output representation. 
We then elaborate on each of the three main building blocks of our MT-DTI model,
the character-embedded Transformer layers (Molecule Transformers, Figure~\ref{fig:moleule_transformer}, Section~\ref{ssec:transformers}), the
character-embedded Protein CNN layers (Protein CNNs, Figure~\ref{fig:protein_cnn}, Section~\ref{ssec:cnns}),
and the dense layers to model interactions between a drug and a protein (Interaction Denses, Figure~\ref{fig:interaction_dense}, Section~\ref{ssec:denses}).
Then, we explain the process for pre-training the molecule transformers (MT) (Section~\ref{ssec:transformers}).
%


\subsection{Model Architecture}
\label{ssec:model_architecture}

The MT-DTI model takes two inputs: a molecule represented by the SMILES~\citep{weininger1988smiles} sequence 
and a protein represented by the FASTA~\citep{lipman1985rapid} sequence.
A molecule represented using the SMILES sequence is comprised of characters representing atoms or structure indicators. 
Mathematically, a molecule is represented as $I_{M}=\{ m_{1}, m_{2}, \dots, m_{L_M} \}$,
where $m_i$ could be either an atom or a structure indicator, and $L_M$ is the sequence length, which varies depending on a molecule.
This molecule sequence is fed into the Molecule Transformers (Section~\ref{ssec:transformers}), to produce a molecule encoding, $M_{enc} \in \mathbb{R}^{E_M}$.
Another type of input, a protein with FASTA sequence, also consists of characters of various amino acids.
A formal protein representation is $I_{P}=\{ p_{1}, p_{2}, \dots, p_{L_P} \}$,
where $p_j$ is one of the amino acids, and $L_P$ is the sequence length, which varies depending on a protein.
This protein sequence is the input of the Protein CNNs (Section~\ref{ssec:cnns}) and generates a protein encoding, $P_{enc} \in \mathbb{R}^{E_P}$.
Note that the encoding vector dimension $E_M$ and $E_P$ are model parameters.
Both of the encodings, $M_{enc}$ and $P_{enc}$ are together fed into the multi-layered feed-forward network, Interaction Denses (Section~\ref{ssec:denses}), 
followed by the last regression layer, which predicts the binding affinity scores.

\subsection{Molecule Transformers}
\label{ssec:transformers}
Molecule Transformers (Figure~\ref{fig:moleule_transformer}) are multi-layered bidirectional Transformer encoders based on the original Transformer model~\citep{vaswani2017attention}.
The Transformer can model a sequence by itself without using a recurrent neural network (RNN) or CNN. 
Unlike these previous sequence processing layers (RNN or CNN), Transformer can effectively encode the relationship among long-distance tokens (atoms) in a sequence.
This powerful context modeling enables many Transformer-based NLP models to outperform previous methods in many benchmarks~\citep{vaswani2017attention,devlin2018bert}.
Molecule Transformers is a modification of the existing Transformer, BERT~\citep{devlin2018bert}, to better represent a molecule
by changing the cost function. 
Before plugging it into the proposed model (Figure~\ref{fig:overall}), 
we pre-train it using the modified masked language model task, which was introduced in the BERT model~\citep{devlin2018bert}.
Each Transformer block consists of a self-attention layer and feedforward layer, 
and it takes embedding vectors as an input.
Therefore the first Transformer block needs to convert an input sequence into the form of vectors using the input embedding.

\subsubsection{Input Embedding}
\label{sssec:input_emnedding}

\begin{figure}
	\centering
	\begin{minipage}[b]{0.8\textwidth}
		\includegraphics[width=\textwidth]{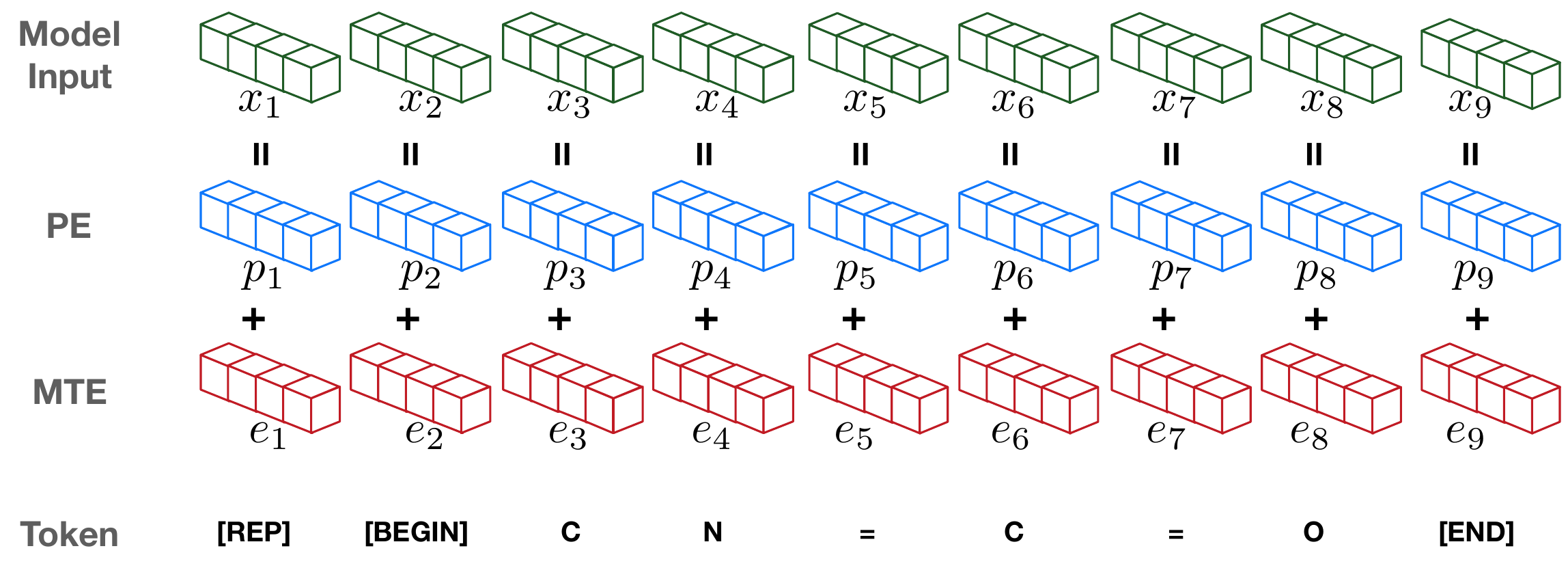}
		\caption{An example of molecule token embedding (MTE) and positional embedding (PE) to make the model input $x_i$ 
		for a given molecule sequence of methyl isocyanate (CN=C=O).
		}
		\label{fig:emb}
	\end{minipage}
\end{figure}

The input to the Molecule Transformers is the sum of the token embeddings and the position embeddings.
The token embeddings are similar to the word embeddings~\citep{mikolov2013distributed}, in that
each token, $m_{i}$ is represented by a molecule token embedding (MTE) vector, $e_i$. 
These vectors are stored in a trainable weights $\text{MTE} \in \mathbb{R}^{V_M \times D_M}$, where $V_M$ is the size of the SMILES vocabulary 
and $D_M$ is the molecule embedding size.
A MTE vector itself is not sufficient to represent a molecule sequence with a self-attention network, 
because a self-attention doesn't consider the sequence order when calculating the attentions, unlike other attention mechanisms.
Therefore, we add a trainable positional embedding (PE)\footnote{Please refer to~\citep{devlin2018bert} for more details.}, $p_{i} \in \mathbb{R}^{L_M^{max} \times D_M}$, to $e_i$ that makes the final input representation, $x_i$ where $L_M^{max}$ is the maximum length of a molecule sequence, which is set to 100 in this study. 
This process is illustrated in Figure~\ref{fig:emb}.

We add five special tokens to the SMILES vocabulary to make a raw molecule sequence compatible with our model. 
[PAD] is for dummy padding to ensure the sequence has a fixed length.
[REP] is a representation token that is used when fine-tuning the Transformer in the proposed MT-DTI model.
[BEGIN]/[END] indicate the beginning or end of the sequence.
This token is useful for the model when dealing with a sequence longer than $L_M^{max}$.
When it is truncated on both sides, the absence of [BEGIN]/[END] tokens will serve as an effective indicator of a truncation.
Methyl isocyanate (CN=C=O), for example, can be represented with 9 tokens;
$$ \text{[REP]} \; \text{[BEGIN]} \; \text{C} \; \text{N} \; \text{=} \; \text{C} \; \text{=} \; \text{O} \; \text{[END]} $$
Each token is transformed into a corresponding vector by referencing $\text{MTE}$ and $\text{PE}$.

\subsubsection{Self-Attention Layer}
\label{sssec:self_attention_layer}
\begin{figure}
	\begin{minipage}[b]{0.28\textwidth}
		\includegraphics[width=0.95\textwidth]{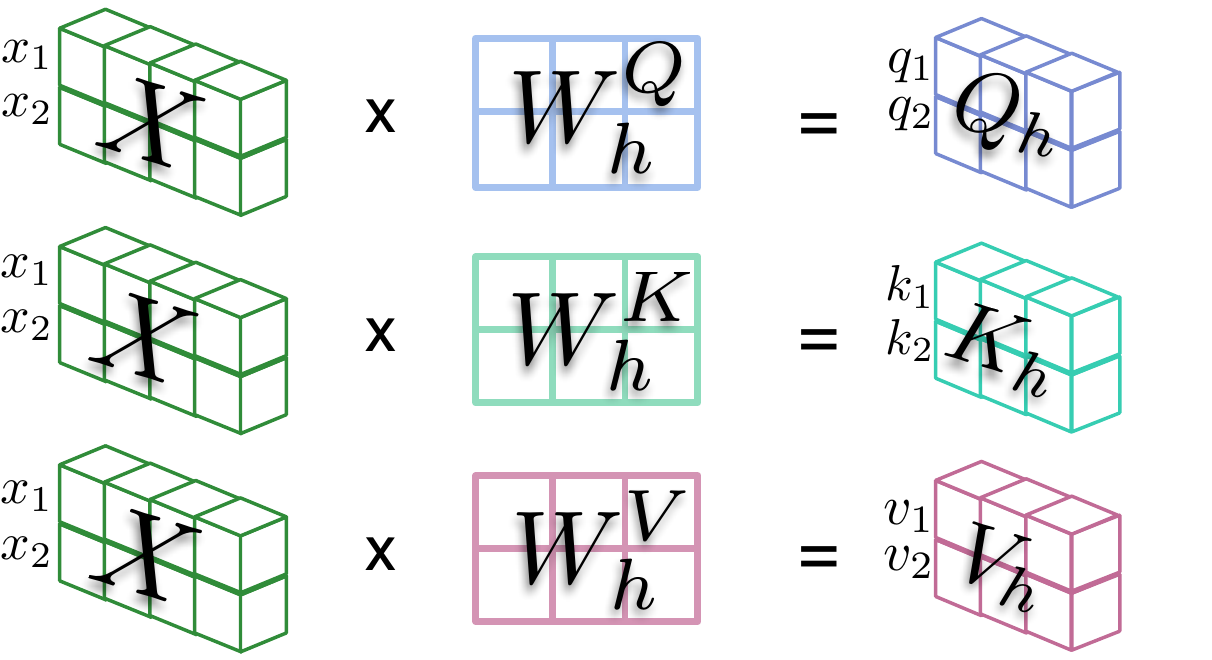}
        \caption{ Query, Key, and Value.}
		\label{fig:att_qkv}
	\end{minipage}
	\begin{minipage}[b]{0.375\textwidth}
		\includegraphics[width=0.95\textwidth]{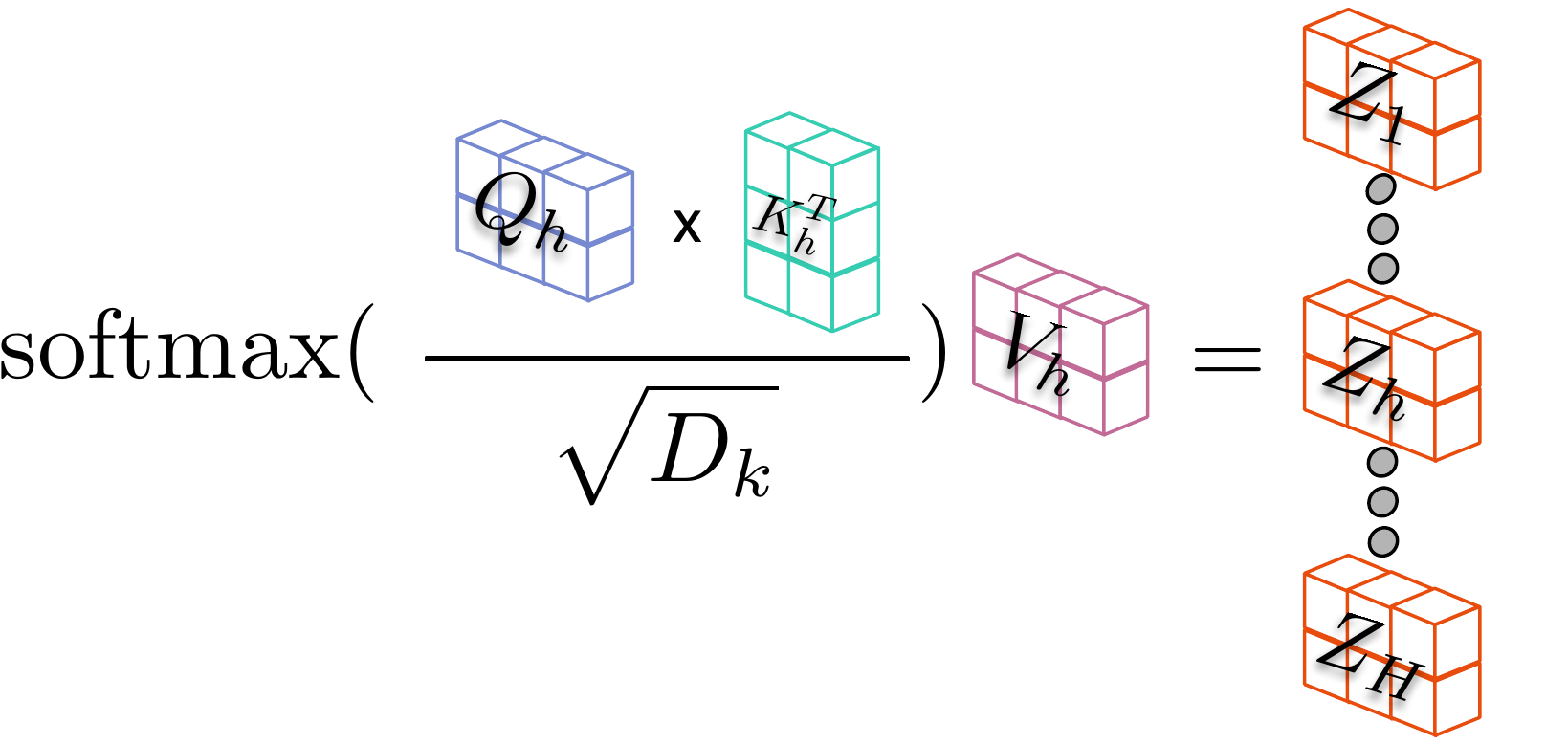}
		\caption{Self-Attention Heads.}
		\label{fig:att_sm}
	\end{minipage}
	\begin{minipage}[b]{0.33\textwidth}
		\includegraphics[width=0.95\textwidth]{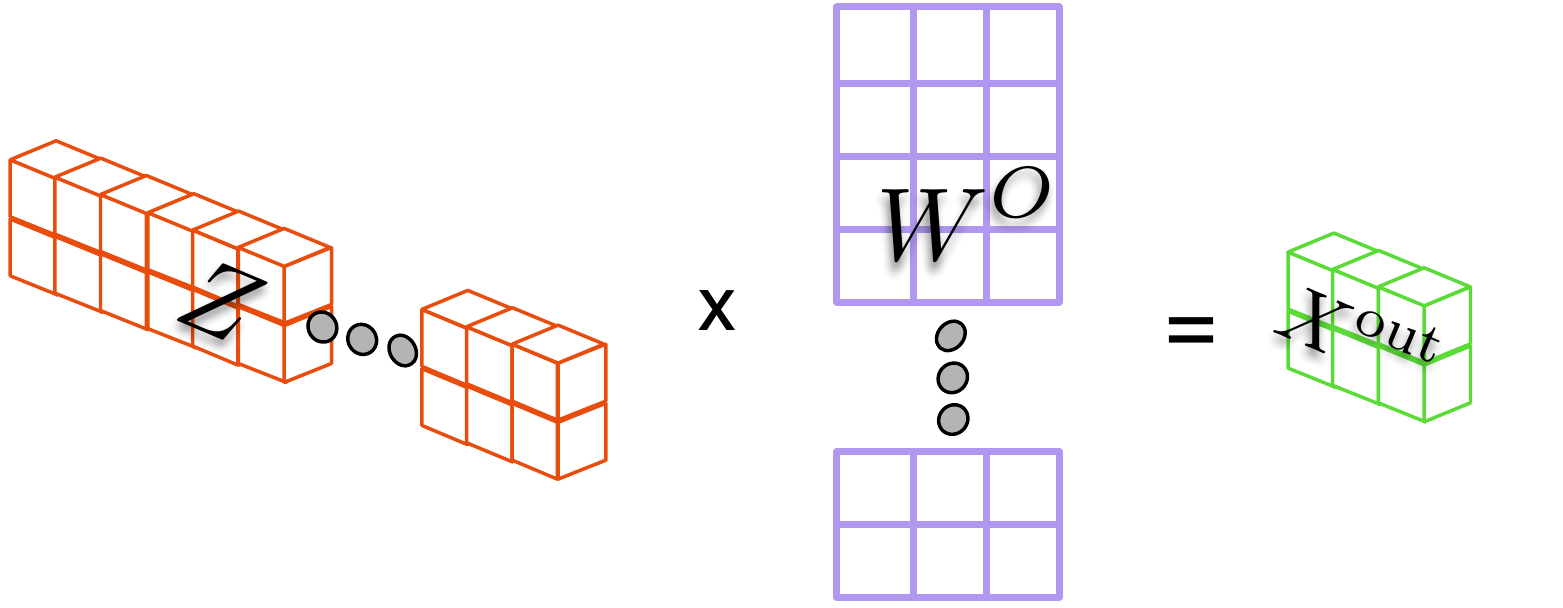}
		\caption{Projected Output.}
		\label{fig:att_ffnn}
	\end{minipage}
\end{figure}

These transformed input vectors, $x_i$, are now compatible with an input to a self-attention layer.
Each self-attention layer is controlled by a query vector ($q_{i}$), key vector ($k_{i}$), 
and value vector ($v_{i}$),
where $i\in \{0,1,\dots, L^{max}_M \} $, all of which are different projections of the input, $X$ ($x_i \in \mathbb{R}^{L^{max}_M \times D_M}$), using
trainable weights, $W^Q \in \mathbb{R}^{D_M \times D_q}$, $W^K \in \mathbb{R}^{D_M \times D_k}$, and $W^V \in \mathbb{R}^{D_M \times D_v}$, shown correspondingly in Figure~\ref{fig:att_qkv}.
Then, the attention weights are computed as:
$$ Z = \text{Attention}(Q, K, V ) = \text{softmax}\left( \frac{QK^T}{\sqrt{D_k}} V \right) \in \mathbb{R}^{L^{max}_M \times D_v}$$
$D_k$ is the dimension of the key (one of the $Z$'s in Figure~\ref{fig:att_sm}).
Thus, the learned relationship between the atoms can span the entire sequence via the self-attention weights.

\subsubsection{Feed-Forward Layer}
\label{sssec:attn_feed_forward_layer}
Similar to multiple filters in convolutional networks, a Transformer can have
multiple attention weights, called multi-head attention.
If one model has $H$-head attention, then it will have $Z_h = \text{Attention}(XW^Q_h, XW^K_h, XW^V_h )$, 
where $h \in \{1,2, \dots, H\}$.
These H number of attention matrices, $Z_h$, are then concatenated (shown on the left of Figure~\ref{fig:att_ffnn}) and 
projected using $W^O \in \mathbb{R}^{H\cdot D_v \times D_M}$ (shown on the middle of Figure~\ref{fig:att_ffnn}) to form a final output of a Transformer, 
$X^{out} \in \mathbb{R}^{L^{max}_M \times D_M}$ (shown on the right of Figure~\ref{fig:att_ffnn}).

\subsubsection{pre-training}
\label{sssec:pre-training}
We adopt one of the pre-training tasks of BERT~\citep{devlin2018bert}, the Masked Language Model.
Since the structure of molecule sequences are shown to be very similar to the structure of 
natural language 
sentences~\citep{jastrzkebski2016learning}
and there are abundant training examples, we hypothesize that predicting masked tokens is an effective way of learning a chemical structure. 
We adopt a special token, [MASK], for this task.
It replaces a small portion of tokens so that the task of the pre-training model is to predict the original tokens.
We choose 15\% of SMILES tokens at random for each molecule sequence, 
and replace the chosen token with one of the special tokens, [MASK] with the probability of 0.8.
For the other 20\% of the time, we replace the chosen token with a random SMILES token\footnote{Since the [MASK] token does not exist when testing, we need to occasionally feed irrelevant tokens when training. } or 
preserve the chosen token, with an equal probability, respectively.
The target label of the task is the chosen token with the index.
For example, one possible prediction task for Methyl isocyanate (CN=C=O) is 
$$ \text{input}: \text{[REP]} \; \text{[BEGIN]} \; \text{C} \; \text{N} \; \text{=} \; \text{[MASK]} \; \text{=} \; \text{O} \; \text{[END]} $$
$$ \text{label}: (\text{C}, 5) $$

\subsubsection{Fine-tuning}
\label{sssec:fine-tuning}
The weights of the pre-trained Transformers (Section~\ref{sssec:pre-training}) are used to initialize the 
Molecule Transformers in the proposed MT-DTI model (Figure~\ref{fig:overall}).
The output of the Transformers is a set of vectors, where the size is equivalent to the number of tokens.
To obtain a molecule representation with a fixed length vector, 
we utilize the vector of the special token, [REP] in the final layer.
This vector conveys the comprehensive bidirectional encoding information for a given molecule sequence, denoted as $M^{rep} \in \mathbb{R}^{D_M}$.

\subsection{Protein CNNs}
\label{ssec:cnns}
Another type of input to the proposed MT-DTI model is a protein sequence.
We modified the protein feature extraction module introduced by~\citep{ozturk2018deepdta} by adding an embedding layer for the input.\footnote{Adding an embedding layer slightly improves the accuracy of the DTI model.}
It consists of 
multi-layer CNNs with an embedding layer to make a sparse protein sequence continuous,
and a pooling layer to represent a protein as a fixed size vector.
For a given protein sequence, $I_p$, each protein token, $p_j$ is converted to a protein embedding vector 
by referencing trainable weights, $\text{PTE} \in \mathbb{R}^{ V_P \times D_P}$, where $V_P$ is the size of the FASTA vocabulary 
and $D_P$ is the protein embedding size.
Let $P \in \mathbb{R}^{L^{max}_P \times D_P}$ be a matrix representing the input protein, 
where $L^{max}_P$ is the maximum length of a protein sequence, which is set to 1000 in this study.
This protein matrix $P$ is fed into the first convolutional layer and convolved 
by the weights $c_1 \in \mathbb{R}^{s_1 \times D_P}$, where $s_1$ is the length of the filter.
This operation is repeated $m_1$ times with the same filter length.
Then this first convolution layer produces a vector $PC_1 \in \mathbb{R}^{L^{max}_P - s_1+1}$, where elements in $PC_1$ convey the $s_1$-gram features across the sequence. 
Multiple convolutional layers can be stacked on top of the previous output of the convolutional layer.
After $v$ number of convolution layers, the final vector, $PC_1 \in \mathbb{R}^{(L^{max}_P - s_1 - s_2  \dots  - s_v +v) \times m_v}$, 
is fed into the max pooling layer.
This max pooling layer selects the most salient features from the vectors produced by the filters from the last layer. 
Then, the output of this max pooling layer is a vector $P^{rep} \in \mathbb{R}^{D_P}$ ($m_v = D_P$). 

\subsection{Interaction Denses}
\label{ssec:denses}
A molecule representation ($M^{rep} \in \mathbb{R}^{D_M}$, Section~\ref{sssec:fine-tuning}) and 
a protein representation ($P^{rep} \in \mathbb{R}^{D_P}$, Section ~\ref{ssec:cnns}) are concatenated to create
the input of Interaction Denses, $MP^{rep} \in \mathbb{R}^{D_M+D_P}$.
Interaction Denses approximates the affinity score through 
a multi-layered feed-forward network with dropout regularization.
The final layer is a regression layer associated with the regression task for the proposed MT-DTI model.
The weights of the network are then optimized according to the mean square error between the network output ($\hat{y}$) and actual affinity values ($y$).

\section{Experiments} 
\label{sec:experiments}

\subsection{Datasets} 
\label{ssec:datasets}

\subsubsection{Drug-Target Interaction} 
\label{sssec:datasets-dti}

\begin{table}[]
\centering\resizebox{\columnwidth}{!}{
    \begin{tabular}{l||c|c|c||c|c|c}
    Dataset & \# of Compounds & \# of Proteins & \# of Interactions  & TRN   & DEV   & TST   \\ \hline \hline
    DAVIS   & 68              & 442            & 30,056  & 20,037 & 5,009  & 5,010  \\
    KIBA    & 2,111            & 229            & 118,254 & 78,836 & 19,709 & 19,709
    \end{tabular}
}
\caption{Statistics of the Davis and Kiba datasets. TRN/DEV/TST:
training, development, evaluation sets.
}
\label{tbl:stat}
\end{table}

The proposed MT-DTI model is evaluated on two benchmarks, Kiba~\citep{tang2014making} and Davis~\citep{davis2011comprehensive},
because they have been used for evaluation in previous drug-target interaction studies~\citep{pahikkala2014toward, he2017simboost, ozturk2018deepdta}.
Davis is a dataset comprised of large-scale biochemical selectivity assays for clinically relevant kinase inhibitors with their respective dissociation constant ($K_d$) values.
The original $K_d$ values are transformed into log space, $pK_d$, for numerical stability,
as suggested by~\citep{he2017simboost} as follows:
$$pK_d = -\log_{10}(\frac{k_d}{1e9}) $$
While Davis measures a bioactivity from one source of score, $K_d$, 
Kiba combines heterogeneous scores, $K_i$, $K_d$ and $IC_{50}$
by optimizing consistency among them.
SimBoost~\citep{he2017simboost} filtered out proteins and compounds with less than 10 interactions for computational efficiency,
and we follow this procedure for a fair comparison.
The number of compounds, proteins and interactions of the two datasets are summarized in Table~\ref{tbl:stat}.
To facilitate comparison and reproducibility,
we followed the same 5-fold cross validation sets 
with a held-out test set which is publicly available\footnote{\url{https://github.com/hkmztrk/DeepDTA/}}.


\subsubsection{pre-training Dataset} 
\label{sssec:datasets-pre}

We downloaded the chemical compound information from the PubChem database~\citep{10.1093/nar/gky1033}\footnote{\url{ftp://ftp.ncbi.nlm.nih.gov/pubchem/Compound/CURRENT-Full/SDF/}}.  
Only canonical SMILES information were used to maintain consistency of representation.
\joyce{Can you just provide 1-sentence why canonical smiles is used instead of other versions?}
A total of 97,092,853 molecules are available in the canonical SMILES format. 


\subsubsection{Drugbank Database} 
\label{sssec:datasets-case}
The DrugBank database comprises a bioinformatics and cheminformatics resource 
that provides known drug-target interaction pairs.
To prove the effectiveness of drug candidates generated by our model, 
we designed a case study (Section ~\ref{ssec:drug_candidates}) using this database.
We extracted 1,794 drugs from the database, excluding any compounds that were used when training our model.
These selected compounds were the input to the trained model (by Kiba dataset) along with a specific protein
to generate corresponding Kiba scores.
The scores were used to find the best candidate drugs targeting that protein.


\subsection{Training Details} 
\label{ssec:training_details}
Molecule Transformer is first trained with the collected compounds from the PubChem database (Section~\ref{sssec:datasets-pre}), 
and then the trained Transformer is plugged into the MT-DTI model for fine-tuning.

\subsubsection{pre-training}
\label{sssec:pre-train}
We use 97 million molecules for pre-training.
Before feeding it to the Molecule Transformer, we tokenize each molecule at the character level.
If the length of the molecules is more than 100, we truncate its head and tail together to have a fixed size of 100. 
We choose the middle part of the longer sequence so that the model can easily distinguish truncated sequences by simply looking at the existence of [BEGIN] and [END] tokens.
The network structure of the Molecule Transformer is as follows.
The number of layers is 8, the number of heads is 8, the hidden vector size is 128, 
the intermediate vector size is 512, the drop-out rate is 0.1, and the activation is Gelu~\citep{hendrycks2016bridging}. 
These parameters are picked from preliminary experiments and the hyperparameters used in the NLP model, BERT~\citep{devlin2018bert}.
We hypothesized that finding a chemical structure might be roughly 2-4 times easier task than finding a language model, 
because the size of the SMILES vocabulary is smaller than natural languages (70 vs 30k).
Although the SMILES vocabulary is 400 times simpler, the number of tokens in the PubChem molecule datasets is about 2.4 times more than what BERT used to pre-train (8B vs 3.3B).
This indicated that the molecules might have more complexity than expected when only considering the size of the vocabulary.
Therefore we used parameters that were 2-4 times smaller than BERT.
We note that there may be other parameter sets that can yield even better performance.
We use the batch size of 512 and the maximum token size of 100, which enables it to process 50K tokens in one batch.
Considering the average length of the compound sequence is around 80, there are approximately 8 billion tokens in the training corpus.
We pre-train Molecule Transformer for 6.4M steps, which is equivalent to 40 epochs (8B/50K*40=6.4M).
With an 8-core TPU machine, the pre-training took about 58 hours.
The final accuracy of the Masked LM task was about 0.9727, which is comparable to the 0.9855 achieved by BERT on natural language.

\subsubsection{fine-tuning}
\label{sssec:fine-tuning_exp}
The specifications of the Molecule Transformer in the MT-DTI model are the same as the one used when pre-training (Section~\ref{sssec:pre-train}).
The Protein CNNs (Section~\ref{ssec:cnns}) consists of 
one embedding layer, three CNN layers, and one max pooling layer.
It uses  128-dimensional vectors for the embedding layer.
For CNN blocks,
we denote the filter size as $K$ and the number of the filter as $L$.
The final model parameter settings of CNNs are $K_1,K_2,K_3=12 (\text{Kiba}), 8(\text{Davis})$ and $L_1=32, L_2=64, L_3=96$.
The max pooling layer selects the best token representations from the last CNN layer,
which makes the feature length as 96.
Interaction Dense (Section~\ref{ssec:denses}) is comprised of three feed-forward layers
and one regression layer.
The layer sizes, when training Kiba, are 1024, 1024, 512 in order of the feature input to the regression layer and the learning rate, $\gamma$, is 0.0001. 
We reduced the network complexity when training Davis due to the small number of training samples.
We use two feed-forward layers of sizes 1024 and 512.
The learning rate is adjusted to 0.001.
The entire network uses the same dropout rate of 0.1.
All the hyper-parameters are tuned based on the lowest mean square error of the development sets for each fold, and the final score is evaluated on the held-out test set with the model at 1000 epochs.

\subsection{Evaluation Metrics} 
\label{ssec:evaluation_metrics}
We use four metrics to evaluate the proposed model:
mean squared Error (MSE), concordance index (CI)~\citep{gonen2005concordance}, $r_m^2$, and area under the precision-recall curve (AUPR). MSE is a typical loss in the optimizer. CI is the probability that the predicted scores of two randomly chosen drug-target pairs, $y_i$ and $y_j$, are in the correct order: 
$$ \text{CI}= \frac{1}{N} \sum_{y_i>y_j} h(\hat{y_i}>\hat{y_j}),$$
where $N$ is a normalization constant (the number of data pairs) and $h(\cdot)$ is a step function~\citep{ozturk2018deepdta}:
\begin{equation*}
    h(x) =
    \begin{cases}
      1, & x>0 \\
      0.5  & x=0 \\
      0, & \text{else}
    \end{cases}
\end{equation*}
The $r_m^2$~\citep{pratim2009two, roy2013some} index is a metric for quantitative structure-activity relationship models (QSAR models).
Mathematically,
$$r_m^2 = r^2 * (1-\sqrt{r^2 - r^2_0}), $$
where $r^2$ and $r^2_0$ are the squared correlation coefficients with and without intercept, respectively.
An acceptable model should produce an $r_m^2$ value greater than 0.5.
Since AUPR is a metric for binary classification, we transform the regression scores to binary labels using known threshold values~\citep{he2017simboost, tang2014making}.
For Davis, pairs with $pK_d \ge 7$ are marked as binding (1), others as no binding (0), 
and for Kiba, pairs with $\text{KIBA score} \ge 12.1$ are marked as binding (1), others as no binding (0).

\subsection{Baselines} 
\label{ssec:baselines}

For the baseline methods, two similarity-based models and one deep learning-based model, the current SOTA,
are tested.
One of the similarity-based models is KronRLS~\citep{pahikkala2014toward},
whose goal is to minimize a typical squared error loss function with
a special regularization term.
The regularization term is given as a norm of the prediction model, 
which is associated with a symmetric similarity measure.
Another similarity-based model is Simboost~\citep{he2017simboost}, 
which is based on a gradient boosting machine.
Simboost utilizes many kinds of engineered features, 
such as network metrics, neighbor statistics, PageRank~\citep{page1999pagerank} scores, and latent vectors from matrix factorization.
The last one is a deep learning model, which is 
the SOTA method in predicting drug-target interactions, called DeepDTA~\citep{ozturk2018deepdta}. 
It is an end-to-end model that takes a pair of sequences, (molecule, protein),
and directly predicts affinity scores from the model. 
Features are automatically captured through back propagation of the multi-layered convolutional neural networks.

\subsection{Results} 
\label{ssec:results}


\begin{table}[]
\centering\resizebox{0.8\columnwidth}{!}{
    \begin{tabular}{c|c||llll}
    Datasets & Method   & CI (std)       & MSE   & $r_m^2$ (std) & AUPR (std) \\ \hline \hline 
    \multirow{4}{*}{Kiba} & KronRLS  & 0.782 (0.001) & 0.411 & 0.342 (0.001)              & 0.635 (0.004) \\
      & SimBoost & 0.836 (0.001)  & 0.222 & \cellcolor{gray!10}0.629 (0.007)              & 0.760 (0.003) \\
      & DeepDTA   & \cellcolor{gray!30}0.863 (0.002)                            & \cellcolor{gray!30}0.194                           & \cellcolor{gray!20}0.673 (0.009)                           & \cellcolor{gray!10}0.788 (0.004)                           \\
      & MT-DTI$^{w/o FT}$    & \cellcolor{gray!10}0.844 (0.001)                             & \cellcolor{gray!10}0.220                            & 0.584 (0.002)                           & \cellcolor{gray!30}0.789 (0.004)                           \\
      & MT-DTI   & \cellcolor{gray!50}\textbf{0.882(0.001)}   & \cellcolor{gray!50}\textbf{0.152}  & \cellcolor{gray!50}\textbf{0.738(0.006)}   & \cellcolor{gray!50}\textbf{0.837(0.003)} \\ \hline
    \multirow{4}{*}{Davis} & KronRLS  & 0.871 (0.001) & 0.379 & 0.407 (0.005)          & 0.661 (0.010) \\
      & SimBoost & 0.872 (0.002)  & 0.282 & \cellcolor{gray!30}0.644 (0.006)          & 0.\cellcolor{gray!10}709 (0.008)  \\
      & DeepDTA   & \cellcolor{gray!30}0.878 (0.004)                            & \cellcolor{gray!30}0.261                           & 0.630 (0.017)                           & \cellcolor{gray!30}0.714 (0.010)                           \\
      & MT-DTI$^{w/o FT}$    & \cellcolor{gray!10}0.875 (0.001)                            & \cellcolor{gray!10}0.268                           & \cellcolor{gray!10}0.633 (0.013)                           & 0.700 (0.011)                           \\
      & MT-DTI   & \cellcolor{gray!50}\textbf{0.887(0.003)}   & \cellcolor{gray!50}\textbf{0.245} & \cellcolor{gray!50}\textbf{0.665(0.014)}      & \cellcolor{gray!50}\textbf{0.730(0.014)}
    \end{tabular}
}
\caption{Test set results of the proposed MT-DTI model, MT-DTI model without fine-tuning (denoted as MT-DTI$^{^w/o FT}$), and other existing approaches.}
\label{tbl:result}
\end{table}

The comparisons of our proposed MT-DTI model to the previous approaches are shown in Table~\ref{tbl:result}.
Reported scores are measured on the held-out test set using 
five models trained with the five different training sets.
The best model parameters are selected based on the development set scores.
MT-DTI outperforms all the other methods in all of the four metrics.
The performance improvement is more noticeable when
when there are many training data where the improvements of Kiba are 0.019, 0.042, 0.065, and 0.04 compared with Davis's improvements of 0.009, 0.016, 0.035, and 0.016, for CI, MSE, $r_m^2$, and AUPR, respectively.
Furthermore, our model tends to be more stable with a larger training set, with the lowest standard deviation for CI and AUPR.
Another interesting point is that our method without fine-tuning (MT-DTI$^{w/o FT}$ in Table~\ref{tbl:result})
produced competitive results.
It outperforms the similarity based metrics and performs better than Deep-DTA for some metrics. 
This suggests that the molecule representation using pre-training learns some useful chemical structure that can be exploited by the interaction denses model. 




\section{Case Study} 
\label{sec:case_study}
We performed a case study using actual FDA-approved drugs targeting a specific protein, epidermal growth factor receptor (EGFR).
This protein is chosen because this is one of the famous genes related to many cancer types. 
We calculated the interaction scores between EGFR and the 1,794 selected molecules based on the DrugBank database (see Section~\ref{sssec:datasets-case} for the details).
These scores are sorted in descending order and summarized in Table~\ref{tbl:drugbank}.

EGFR is a transmembrane protein that is activated by binding of ligands such as epidermal growth factor (EGF) and transforming growth factor alpha (TGFa) \citep{herbst2004review}. Mutations in the coding regions of the EGFR gene are associated with many cancers, including lung adenocarcinoma \citep{sigismund2018emerging}. Several tyrosine kinase inhibitors (TKIs) have been developed for the EGFR protein, including gefitinib, erlotinib, and afatinib. More recently, Osimertinib was developed as a third generation TKI targeting the T790M mutation in the exon of the EGFR gene \citep{soria2018osimertinib}. Since the direct binding of these drugs to EGFR protein is well known, we tested whether our proposed model can identify known drugs for the EGFR protein. 

\subsection{Biological Insights} 
\label{ssec:drug_candidates}

The result indicated that our model successfully identified known EGFR targeted drugs as well as novel chemical compounds that were not reported for association with the EGFR protein. For example, the first and second generation TKIs, such as Erlotinib and Gefitinib, and Afatinib, respectively, were predicted to exhibit high affinity to the EGFR protein (Table 3). Lapatinib \citep{medina2008lapatinib}, which inhibits the tyrosine kinase activity associated with two oncogenes, EGFR and HER2/neu (human EGFR type 2), was predicted to exhibit the highest affinity.  Osimertinib was also identified. Interestingly, chemical compounds targeting opioid receptors (naltrexone hydrochloride, nalbuphine hydrochloride, and oxycodone hydrochloride trihydrate) for pain relief, antihistamines (methdilazine hydrochloride and astemizole), antipsychotic medication for schizophrenia (Prolixin Enanthate and Asenapine), and corticosteroids for skin problems (Triamcinolone acetonide sodium phosphate, Oxymetazoline hydrochloride, Desonide) were predicted to be associated with EGFR. Among these chemical compounds, Astemizole was suggested as a promising compound when treated with known drugs for lung cancer patients \citep{ellegaard2016repurposing,de2017combination}. Therefore, further investigations of these chemicals may provide a new therapeutic strategy for lung cancer patients. 

\begin{table}[]
\begin{tabular}{llll}
Ranking & Compound ID & Compound Name                            & KIBA Score \\ \hline \hline
1       & 208908      & \textbf{Lapatinib$^*$}                            & 14.002403  \\
2       & 11557040    & \textbf{Lapatinib Ditosylate$^*$}                 & 13.811217  \\
3       & 10184653    & \textbf{Afatinib$^*$}                             & 13.404812  \\
4       & 16147       & Triamcinolone Acetonide Sodium Phosphate & 13.147043  \\
5       & 5485201     & Naltrexone Hydrochloride                 & 13.114577  \\
6       & 123631      & \textbf{Gefitinib$^*$}                            & 13.111686  \\
7       & 60699       & Topotecan Hydrochloride                  & 13.108758  \\
8       & 5360515     & Naltrexone                               & 13.065864  \\
9       & 441351      & Rocuronium Bromide                       & 13.032806  \\
10      & 6918543     & Almitrine Mesylate                       & 13.016999  \\
11      & 176870      & \textbf{Erlotinib$^*$}                            & 12.885199  \\
12      & 23422       & Tubocurarine Chloride Pentahydrate       & 12.87076   \\
13      & 6000        & Tubocurarine                             & 12.809549  \\
14      & 11954379    & \textbf{Erlotinib Variant$^*$ }                   & 12.782704  \\
15      & 11954378    & \textbf{Erlotinib Hydrochloride$^*$}              & 12.768639  \\
16      & 3389        & Prolixin Enanthate                       & 12.737285  \\
17      & 23724988    & Oxycodone Hydrochloride Trihydrate       & 12.709352  \\
18      & 14676       & Methdilazine Hydrochloride               & 12.662965  \\
19      & 5281065     & Ibutilide Fumarate                       & 12.650397  \\
20      & 9869929     & Avanafil                                 & 12.635439  \\
21      & 60700       & Topotecan                                & 12.618897  \\
22      & 5360733     & Nalbuphine Hydrochloride                 & 12.610958  \\
23      & 5282487     & Paroxetine Hydrochloride Hemihydrate     & 12.608804  \\
24      & 66259       & Oxymetazoline Hydrochloride              & 12.557486  \\
25      & 5311066     & Desonide                                 & 12.538858  \\
26      & 2247        & Astemizole                               & 12.536284  \\
27      & 11954293    & Asenapine                                & 12.534941  \\
28      & 11304743    & Riociguat                                & 12.527533  \\
29      & 82153       & Flunisolide                              & 12.527164  \\
30      & 71496458    & \textbf{Osimertinib$^*$ }                         & 12.507524 
\end{tabular}
\caption{Compound ranking based on the predicted Kiba scores when the target is EGFR protein. All compounds are from Drugbank database excluded any compounds in Kiba dataset. \textbf{[Compound Name]$^*$} represents a known EGFR targetting drug.
}
\label{tbl:drugbank}
\end{table}


\section{Related Work} 
Predicting drug-target interaction traditionally focused on a binary 
classification problem~\citep{yamanishi2008prediction,bleakley2009supervised,van2011gaussian,cao2012large,gonen2012predicting,cobanoglu2013predicting,cao2014computational,ozturk2016comparative}.
The most recent approach tackling this binary classification problem is an interpretable deep learning based model~\citep{gao2018interpretable}.
Although these methods show promising results on binary datasets, 
they are simplifying protein-ligand interactions by thresholding affinity values.
In order to model these complex interactions, several methods have been proposed, which can be categorized into three kinds.
The first category of these models is molecular docking~\citep{trott2010autodock, luo2016molecular}, 
which is a simulation-based method. 
These methods are not scalable, due to heavy preprocessing.
To overcome this downside, the second category, similarity-based methods, was proposed.
They are KronRLS~\citep{pahikkala2014toward} and SimBoost~\citep{he2017simboost}, which is based on the calculation of similarity matrix of inputs.
With the advent of deep learning, two deep learning-based methods have been proposed~\citep{gao2018interpretable,ozturk2018deepdta}.
Like these models, our model is also based on deep learning, 
but our proposed model has a better molecule representation, 
and improves its performance through a transfer learning technique.

Deep learning-based transfer learning, pre-training and fine-tuning, have been applied to various tasks such as computer vision~\citep{rothe2015dex,ghifary2016deep}, 
NLP~\citep{howard2018universal}, speech recognition~\citep{jaitly2012application,lu2013speech}, and health-care applications~\citep{shin2017classification}.
The idea is to use appropriate pre-trained weights to improve results in corresponding tasks, 
which also can be found in our experimental results.

\section{Discussion} 
This paper proposes a new molecule representation using the self-attention mechanism, 
which is pre-trained using publicly available big data of compounds.
The trained parameters are transferred to our DTI model (MT-DTI) so that it can be fine-tuned using two DTI benchmark data.
Experimental results show that our model outperforms all other existing methods with respect to four evaluation metrics.
Moreover, the case study of finding drug candidates targeting a cancer protein (EGFR) shows that
our method successfully enlists all of the existing EGFR drugs in top-30 promising candidates.
This suggests our DTI model could potentially yield low-cost drugs and provide personalized medicines.
Our model can be further improved as the proposed attention mechanism is also applied to represent proteins.
However, we didn't explore this direction for two reasons.
One reason is that the length of a protein sequence is ten times longer than a molecule sequence on average, 
which takes a considerable amount of time for computation.
Another reason is the need for a protein dataset which contains enough sufficient information to pre-train the model.
Unfortunately, such dataset is not readily available.


\acks{This work was supported by the National  Science Foundation award IIS-\texttt{\#}1838200, AWS Cloud Credits for Research program, and Google Cloud Platform research credits.}

\bibliography{bdti}



\end{document}